\documentclass[conference]{IEEEtran}
\IEEEoverridecommandlockouts
\usepackage{cite}
\usepackage{amsmath,amssymb,amsfonts}
\usepackage{algorithmic}
\usepackage{graphicx}
\usepackage{textcomp}
\usepackage{xcolor}

\usepackage{times,amsmath,epsfig,amssymb,cite,bm,float,epstopdf,graphicx,graphics,amsthm}
\usepackage{balance,multicol}
\usepackage{color}
\usepackage[stable]{footmisc}
\usepackage{amsthm}
\usepackage{pdfsync}
\usepackage{subcaption}
\usepackage{amssymb}
\usepackage{amsmath}
\usepackage{ifthen}
\usepackage{cool,booktabs}
\usepackage{makeidx}
\usepackage{verbatim}
\usepackage[ruled,vlined]{algorithm2e}
\usepackage[font=small]{caption}

\def\BibTeX{{\rm B\kern-.05em{\sc i\kern-.025em b}\kern-.08em
    T\kern-.1667em\lower.7ex\hbox{E}\kern-.125emX}}
\begin{document}

\title{Reinforcement Learning-based Joint Path and Energy Optimization of Cellular-Connected Unmanned Aerial Vehicles\\
\thanks{Many thanks to Eng. G.A. Houshmand as well as KTH Library RIN 7655 for their financial support.}
}

\author{\IEEEauthorblockN{Arash Hooshmand}
\IEEEauthorblockA{\textit{School of Electrical Engineering and Computer Science} \\
\textit{KTH Royal Institute of Technology}\\
Stockholm, Sweden \\
hooshmand@kth.se}
}

\maketitle

\begin{abstract}
Unmanned Aerial Vehicles (UAVs) have attracted considerable research interest recently. Especially when it comes to the realm of Internet of Things, the UAVs with Internet connectivity are one of the main demands. Furthermore, the energy constraint i.e. battery limit is a bottle-neck of the UAVs that can limit their applications. We try to address and solve the energy problem. Therefore, a path planning method for a cellular-connected UAV is proposed that will enable the UAV to plan its path in an area much larger than its battery range by getting recharged in certain positions equipped with power stations (PSs). In addition to the energy constraint, there are also no-fly zones; for example, due to Air to Air (A2A) and Air to Ground (A2G) interference or for lack of necessary connectivity that impose extra constraints in the trajectory optimization of the UAV. No-fly zones determine the infeasible areas that should be avoided. We have used a reinforcement learning (RL) hierarchically to extend typical short-range path planners to consider battery recharge and solve the problem of UAVs in long missions. The problem is simulated for the UAV that flies over a large area, and Q-learning algorithm could enable the UAV to find the optimal path and recharge policy.
\end{abstract}


\section{Introduction}
\label{sec:introduction}
The ambitious goal of connectivity anywhere includes the connectivity in the sky as an important part of it. Providing unmanned aerial vehicles (UAVs) and other aircraft with the wireless connectivity can be realizable with direct air-to-ground communications (DA2GC) \cite{b1}. DA2GC is a promising solution to achieve the goal, and are superior to satellite communications due to their lower latency and higher throughput capabilities \cite{b1}. The aim is to integrate UAVs into the cellular networks, which will then be called cellular-connected UAVs \cite{b2}. Although cellular networks provide advantages such as ubiquitous connectivity of UAVs, interference due to uplink (from UAVs to base stations (BSs)) and downlink (from BSs to UAVs) communication becomes a challenge \cite{b3,b4, b5}. BSs include both ground BSs (GBSs) and Aerial BSs (ABSs) \cite{b5, b6, b7}. Furthermore, UAVs such as the one showin in Fig. 1 need to have great flexibility but face certain challenges such as avoidance of no-fly zones and collisions. Another challenge is that the battery of UAVs is finite and may not be sufficient to fulfill longer distance missions beyond a certain limit accessible with one full battery. To address these challenges, mobility management becomes important with respect to path planning of UAVs. For example, the authors in \cite{b5} reference to several path planning papers that use dynamic programming to minimize the time for a UAV having a signal to noise ratio (SINR) less than a threshold. Authors in \cite{b8} minimize the path completion time of UAVs with a certain level of SINR constraints by using graph theory and convex optimization, and in \cite{b9} propose a machine learning powered path optimization to minimize the interference to the ground network. Researchers in \cite{b9} and \cite{b10} design a circular trajectory to maximize sum-rate of ground users by considering propulsion energy consumption of UAVs. However, these studies do not consider the battery limitation of UAVs; and those studies such as \cite{b11} that address battery life-time usually do not involve communication systems features including connectivity and interference. 

\begin{figure}[htbp]
\centerline{\includegraphics[width=.47\textwidth]{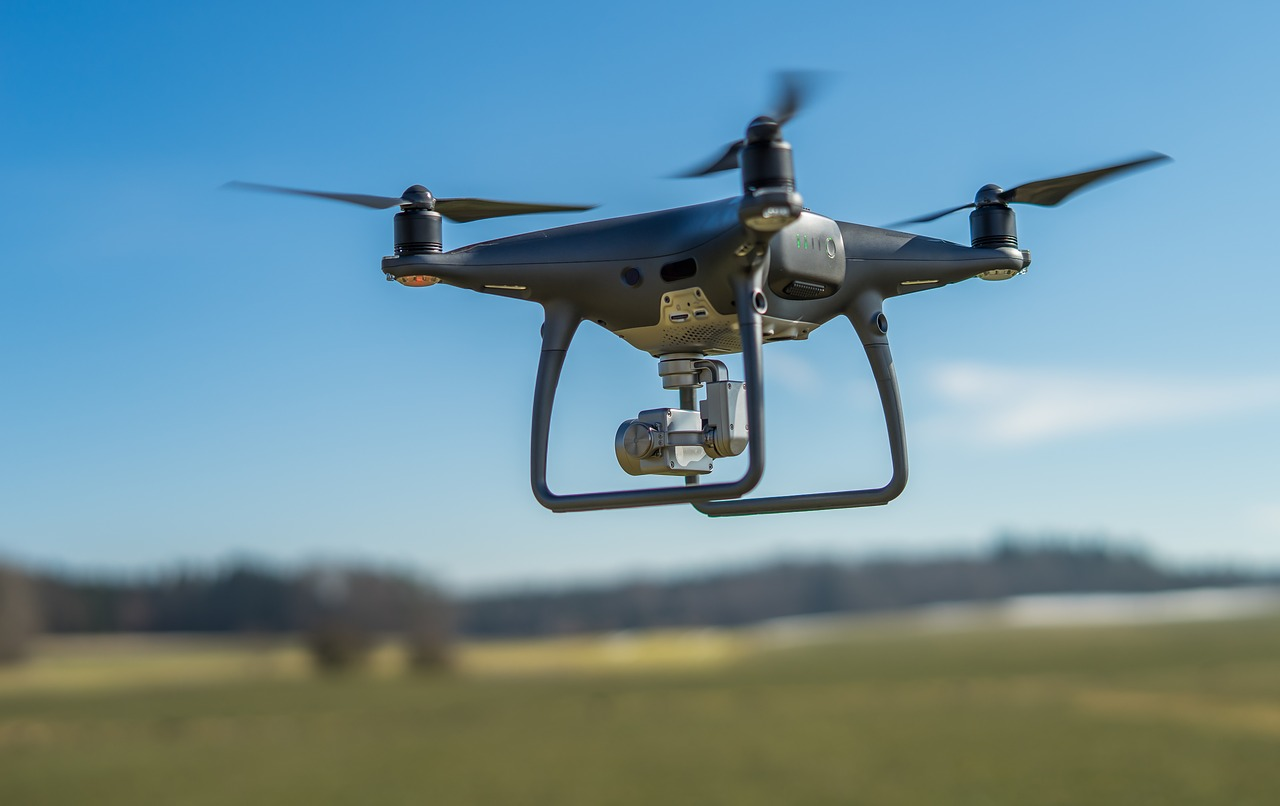}}
\caption{A typical drone flying over rural areas can fly longer and further by battery recharge facilities. The photo with Creative Common License by Lars\_Nilssen in Pixabay 2018. Accessed September 16, 2020 from https://pixabay.com/photos/drone-uav-quadrocopter-hobby-sky-3198324 }
\label{fig:f0}
\end{figure}

\subsection{Background}

Wireless communication systems have their own technical characteristics, strengths and weaknesses. For example, signals attenuate with distance from the transmitter antenna as they propagate in media, and electromagnetic interference (EMI) occurs when in the radio frequency spectrum, external signals disturb the transmission of a signal. To see a simple example model with typical settings, one can refer to the table 1 at \cite{b12} as a basic general introduction, e.g. the  Carrier frequency in mobile communications can be assumed to be around $2$ GHz with $20$ MHz bandwidth which will determine the signal-to-noise ratio and throughput and the information bit rate. 
Consider the uplink communication of our UAV is to continuously deliver videos and data to a center via cellular network BSs. 
For simplicity, one can consider that the UAV operates in a plain rural area without tall buildings or vast water surfaces, such as lakes and seas. Hence, generic models such as the Rician channel and free-space path loss model can be used for the data link between the UAV and its connected BS. 
Let's also focus on a single cellular-connected UAV flying with a constant speed of $V$ meters per second. The set of Ground BSs (GBSs) are denoted by $G = {g_1, g_2, . . . , g_{N_g}}$, with $|G| = N_g$, and also as a more futuristic scenario one can also assume a set of ABSs denoted by $A = {a_1, a_2, . . . , a_{N_a}}$, with $|A| = N_a$. The constraint for the UAV is to have an acceptable cellular connection (i.e., maintaining a link with an information rate above a given threshold) during its trajectory and it is not very different to extract the considerations in relation to GBSs and ABSs because similar Rician models can be used but with different coefficients. 

UAVs use different sensors and systems; for example, for coordination systems, since most of the modern UAVs are equipped with GPS, they use WGS84 coordinate system. Meanwhile, their computer can also use other coordinate systems for practical purposes, simultaneously. For example, for path planning on a grid map, following many researchers and papers such as \cite{b13}, a Cartesian coordinate system is used in which the location of $i_{th}$ BS can be denoted as ($x_i, y_i, z_i$). Similar to \cite{b13}, without affecting the operating area in its general form and only in order to simplify the calculations, it is assumed that all GBSs have the same altitude, which is equal to HG (i.e., $z_i = z_j = HG$ for $\forall i, j \in [1, N_g]$). All ABSs also fly at the same constant altitude, which is equal to HA (i.e., $z_i = z_j = HA$ for $\forall i, j \in [1, N_a]$). This would enable us to deal with a 2D map instead of 3D and leave the altitude planning to the complex fine 3D planners on which our coarse path planner will work and will be clarified later. Furthermore, we denote the start location with $Start = (S0, S1, S2)$ and the end location with $Destination = (D0, D1, D2)$; despite the fact that in dynamic programming-based approaches (including Q-learning approach implemented in this project), the problem will be solved for any Start point, the shortest path is extracted up to the Destination. 
In addition, as the UAV will be moving, we denote its location at time t with $(x(t), y(t), H(t))$. Within the intervals that H(t) does not change or whenever its change is negligible, it can be denoted by its constant value H. 

Each BS and the UAV are assumed to be equipped with a single omnidirectional antenna with unit gain. Moreover, similar to the settings at \cite{b14}, the channel between the UAV and a BS is assumed to be dominated by the LoS link. Given the specifications of the BSs, a range $R$ is defined for minimum SNR needed for the communication between the UAV and BSs. This could be computed using the following equation \cite{b14}: 
\begin {equation}
R=\sqrt{(\gamma_0/S_{min}-(H - HG)^2)}
\end{equation}
where $\gamma_0=P  \beta_0/\sigma^2$ denotes the reference SNR, $P$ is the transmission power of GBS, $\sigma^2$ as the noise power at the receiver, and $\beta_0$ as the channel power gain at the reference distance of 1 m. $S_{min}$ is the minimum required SNR value for the communication between the UAV and the GBSs. The corresponding formulas for ABSs are achievable easily by replacing corresponding values of ABSs instead of those of GBSs. 
The path loss between the UAV and BSs, $\xi$ is given at \cite{b15} as
\begin{equation} 
\xi (dB) = 20 log10(d) + 20 log10(cf) - 147.55 \end{equation} 
where cf is the carrier frequency and d is the Euclidean distance between the UAV and BS and the adjustment constant added is experimentally obtained. Therefore, distance to the BSs is a key factor in the signal strength and consequently in the available information bit rate between the UAV and the corresponding BS given that the minimum bit rate should be guaranteed, there is a maximum distance away from the BS is feasible for the UAV and beyond it is an infeasible no-fly zone unless that point is covered by another BS. In practice, the maximum allowed distance and feasible area is determined by other factors such as weather conditions or the traffic of other UAVs and wireless systems. However, the focus of this work does not include these physical issues and after this brief introduction to the underlying physics as background, we consider the UAV's computer mostly operating at a higher level where the maps of feasible and infeasible areas is made available to the UAV and it learns a suitable route to the Destination. However, in both possible scenarios, whether the map is sent to the UAV from a center or the map is produced by the UAV after receiving external information as well as sensor signals, the UAV is capable of using some of its sensors or even sensor fusion techniques to double check the facts of its environment to be sure the world is as it is supposed to be. The environmental information can be used just as a tool to double check the UAV is at the right position; for example, by checking its distance to the nearest BS to be within a given range. Therefore, we suppose that the UAV knows its own location using sensor fusion, and is in control of its movement. Nevertheless, details of these techniques and technologies is not at the scope of the current article. Our work in this paper is only an additional complementary unit to the outcome of dozens of valuable works, such as \cite{b16} and \cite{b17} done on UAVs path planning and its contribution is to sit on top of the great path planners such as the ones developed in \cite{b12} and \cite{b13} that adjust parameters such as UAV's altitude in order to solve one remaining problem: battery limits. Our product acts as a coarse 2D planner on large area cells in each of which there is a fine 3D planner for small area cells that form each of large area cells. It means that in our additional 2D coarse path planner, we can consider a constant altitude because we are developing a unit on top of the 3D fine path planner that will consider and manage the altitude planning. These notions will be made clear. 



As a bird's-eye view, in this paper, we simulate a system of one UAV similar to Fig. 1 where the UAV can benefit from power stations (PSs) in the network to recharge its battery and perform longer missions. Buildings, trees, and small pools are allowed but most of the area will be considered plain and empty. Optimization of path planning is performed via Q-learning, a popular version of reinforcement learning (RL), \cite{b18} considering energy depletion constraints in addition to no-fly zones defined by wireless connectivity, interference and other factors but received as a 2D map. The reason to use RL is that dynamic programming based on Bellman equation, that is the theoretically correct approach to solve this type of problems, would use a Brute force algorithm looping through all possible states and actions that makes it impractical for complex problems with large action-state spaces. Q-learning is a feasible alternative that usually converges quickly and covers most of statistically important routes; therefore, results in near-optimal solutions. \cite{b19}

In practice, UAVs are often considered as user equipments (UEs) to deliver their videos and other application data in missions such as surveillance. We propose a simple yet effective machine learning (ML) algorithm in which the UAV receives the map of PSs and no-fly zones and finds its optimal path by considering rewards and punishments for favorable PSs and infeasible areas, respectively. In this project, the map focuses on locating PSs as well as no-fly zones rather than BSs because BSs are not only ground BSs (GBSs) but also ABSs \cite{b5, b6, b7}, and all that matters is the feasible and infeasible areas that are given to the UAV as a map. When we consider Flying UAVs, they change their locations; therefore, the map is dynamic and can vary from time to time due to aerial bodies such as ABSs; therefore, the connectivity regions change in the map dynamically. If they do not act as ABSs, they still play a role in interference as well as security issues such as potential accidents, and hence inevitably impose dynamic changes on feasible and infeasible regions. The system can be regarded as an example generic model to show the principles of autonomous agents' artificial intelligence in the future. We do not have a clear view what the future life with many types of UAVs will look like; therefore, we should reserve the cautious measures considering a dynamic map that may change and need updates due to a great variety of reasons; for example, the position of other airborne bodies.

\section{System Model}

The UAV begins from a Start point and ends at a Destination to perform its mission. The UAV also has the network 2D map including the distribution of available PSs and feasible and infeasible zones. Remember that this map is for the coarse planning in which we do not deal with altitude in it, or we sometimes alternatively may consider a fixed altitude for it in simulations, as the altitude would be determined in the 3D fine planning. The 2D grid also consists of same sized square cells described in the Network model below. Hence, the UAV movement is from one cell to the neighboring cell on the map at a time along its trajectory. The model is designed to be scalable and to easily support any number of UAVs traveling between any two cells on the grid. Although in this work, we show the results for one UAV and its achieved potential trajectories i.e. its optimal paths from any cell on the grid to Destination. The reason for showing the route from each cell to Destination is to cover even the worst case scenarios in the stochastic environment in which unpredictable changes can happen; for instance, rare strong winds can relocate the UAV to any distance away from its position. The UAV aims to avoid any no-fly zones while following its optimal path.   

The UAV can land on and take off vertically from cells such as Destination, Start and PSs and the time and energy for these operations are ignored in this simulation for simplicity. We assume that the onboard battery can support only 10 minutes of flying while the speed of the UAV is 800 meters per minute or equivalently 13.3 meters per second. These numbers are arbitrary but realistic with respect to the current models available in the market; however, one can change them to exact parameters of his UAV and run the simulations again to get the results optimized for his own project. \cite{b20} To analyze the system, we focus on the Destination, the number and location of PSs as well as no-fly zones. The UAV can move in 4 directions i.e. right, left, forward and backward. Its movement is considered to be from the center of its current cell to the center of any of its neighboring cells. These simplifications do not reduce the generality of the problem but make it easy to realize the proof of concept by focusing on the main issues. The energy consumed while flying from one cell to the neighboring cell is considered as the constant travel cost. Since the battery charge should always be positive, a UAV with full battery cannot fly away more than 10 cells without recharging at PSs. 

\subsection{Network model}
The environment of our simulation is considered to be an area of 16km*16km. The entire region under study is modeled as a 20*20 virtual grid that means the distance between the centers of each two neighboring grid cells is $800$ meters i.e. each small cell in this paper is the entire grid in typical papers such as \cite{b12} that is itself considered a 20*20 grid in their work for 3D fine path planning that includes determination of the altitude too. This is the meaning of hierarchical RL path planning when we say that our 2D coarse path planer is an extension on top of 3D fine path planners acting as its base. Since the UAV has limited battery, the cells that contain PSs are strategic rewarding cells. PSs are wireless i.e. they do not need cable coupling to transfer the charge thanks to wireless technologies such as RF charging. Moreover, PSs are distributed in different locations in the network to provide a battery recharge service for the UAV. PSs also can have different altitudes that are ignored in the 2D planning. 

In the current simplified version, we assume that available PSs always have enough energy to fully charge the batteries of UAVs. In future work, if it were necessary, one can add the complexity of having PSs that have low charge or are empty at the moment due to recent charging of other UAVs. The size of grid cells can be set flexibly. It can be set based on battery capacity and energy consumption, connectivity, and interference, speed of UAVs and so on. 
The main focus is on energy management; hence, the large grid corresponding to the coarse planning is designed based on the battery energy consumption. Thus, any significant change in the UAV's speed and energy consumption or in battery capacity may require new decisions on grid design in the future scenarios.

\section{Problem Formulation}
The objective is to minimize the mission time or equivalently to reach to the Destination through shortest path. Formulation and optimization of the problem by ML guarantees its scalability and makes it applicable to very large networks covering long distances while there are PSs available in the environment. For example, Bekhti et al. solve a similar problem as a linear programming problem for a scenario with an area size of 250m*250m \cite{b21} but we could solve our problem as a Q-learning problem for a scenario with an area size of 16km*16km. In general, formulation of the problem for large area networks cannot be done using linear or dynamic programming and the solution has to use statistical approaches such as those used in RL. The path optimization problem for the 2D coarse planning can simply be formulated as follows:
\begin{equation}
\text{Minimize} \quad  T
\label{eq:opt}
\end{equation}
subject to 
\begin{equation}
e \geq e\_min \label{eq:energy}
\end{equation}
\begin{equation}
R \geq R_{thr} \label{eq:link_quality}
\end{equation}
\begin{equation} \label{eq:start_battery}
e(S) = E_{max}  
\end{equation}

The objective is to minimize the total time, T, during which the UAV travels from  the start point to the destination point. The UAV can complete long missions by stopping at PSs to recharge its battery to have enough energy and increase its flight time until reaching to the Destination. Mathematically, we can formulate the problem constraints as follows:

The first constraint in (\ref{eq:energy}) asserts that the UAV should always have some energy. $e$ is the remaining battery energy and cannot be negative or even less than  the minimum energy for UAV to land on a PS. However, one may consider airborne PSs that are likely in the future potential scenarios. In this case, the e\_mean can be almost zero. Running out of battery during the flight means the UAV would crush. 

The constraint in (\ref{eq:link_quality}) is to ensure that during the trajectory, the uplink information bit rate of the UAV is greater than the threshold information rate. This constraint is currently true of all feasible cells. However, we keep it it to reserve for updating the next versions of the algorithm which may loosen this condition with a tolerance of limited violation of this condition according to a given constraint such as the one studied in \cite{b13}. 




The constraints in (\ref{eq:start_battery}) dictates that the battery is full at the Start, thus the UAV starts its mission with full battery. 

\begin{algorithm}
Initialize variables and the grid: Set the Q-learning state space $S$ as a 20*20 grid network, set the Start and Destination cells, set the action space $A$ as four main direction movements i.e. Up, Down, Left and Right, set the table of rewards of each cell effectively with respect to the battery charge i.e. PS cells to 1, regular cells to -0.1, Destination cell to 1000, and experimentally optimized, no-fly cells to -30, set the table of values of each cell $V[s]$ with reward values initially but these values will be calculated, corrected and completed during the program run, Initialize state-action values holders i.e. Q and $Q[s]$ as well as the array of changes in Q values i.e. $deltas[]$ to be filled later during the program run, set $Q[s, a]$ with initial values equal to rewards at each corresponding cell or state and obtain the initial action of each state with the movement to the neighboring cell that contains the maximum reward, as well as experimentally chosen discount factor $\gamma = 0.9$, learning rate $\alpha = 0.1$, epsilon for greedy search $eps = 0.9/t$; and $N = 100000$ as the number of epochs...

\For{$i$ in range N}{
increment $t$ (whatever its unit is) \newline
$grid.set\_state(s)$ \newline
$a = argmax(max(Q[s]))$ \newline
$biggest\_change = 0$ \newline
\While{not grid.game\_over()}{
a = random\_action(a, eps=0.9/t) \newline 
r = grid.move(a) \newline 
s2 = grid.current\_state() \newline 
increment totalcount \newline 
$alpha = ALPHA / (0.995 + update\_counts\_sa[s][a] * 0.005)$ \newline 
$update\_counts\_sa[s][a] += 1 $ \newline 
$old\_qsa = Q[s][a]$ \newline 
$a2, max\_q\_s2a2 = max\_dict(Q[s2])$ \newline 
$Q[s][a] = Q[s][a] + alpha*(r + GAMMA*max\_q\_s2a2 - Q[s][a])$ \newline 
$biggest\_change = max(biggest\_change, np.abs(old\_qsa - Q[s][a]))$ \newline 
$update\_counts[s] = update\_counts.get(s,0) + 1$ \newline 
s = s2 \newline 
a = a2

}
deltas.append(biggest\_change)

policy = \{\} \newline
V = \{\} \newline

\For{s in grid.actions.keys()}{
    a, max\_q = max\_dict(Q[s]) \newline
    policy[s] = a \newline
    V[s] = max\_q \newline
}
}
\caption{Q-Learning Algorithm} \label{ho}
\end{algorithm}

\section{Joint Path and Energy Optimization with RL}

We want to keep the UAV connected while directing it to its Destination through feasible cells with controlled interference. RL is suitable for this purpose because it gives the UAVs the flexibility to optimize their path with respect to the constraints. Particularly, Q-learning as a model-free RL is capable of learning optimal policies to tell the agent what actions should be chosen under certain conditions. 

The agent, i.e., the UAV, can fly over environment that is the entire square area modeled as grids shown in Fig. 2. Each grid cell is a state and the UAV can change its state by choosing one of four directional movement options, i.e., Right, Left, Up and Down that form the action set. If the UAV lands on a PS cell, it will receive a positive reward. On the other hand, if it enters no-fly zone cells, it will get a negative reward. Using negative rewards is preferred over pruning them out because it makes the system flexible to be used in a greater range of potential future applications in the future such as the case in \cite{b13} in which violating some conditions is tolerated up to certain constraints when accomplishing the task even with some imperfections is preferred over its termination. For example, it may be preferred to receive lower quality videos over a short distance of the trajectory rather than stopping the work because of limited bandwidth for a short while. The rewards are experimentally adjusted. In this project, the rewards are calculated based on a formula that uses fixed reward values for PSs and no-fly zones adjusted by its distances to critical points such as the Destination. In all other cells which have neither PS cell rewards nor infeasible cell penalties, a small negative cost is applied as the energy consumed in order to make the UAV proceed to the Destination without wasting time visiting unnecessary cells during its trajectory. 

In Q-learning, the total reward at each cell is the sum of its immediate reward plus all of the future rewards discounted using a discount factor less than 1. As seen in Algorithm 1, after initialization of the environment, i.e., the grid as well as a look-up table by assigning rewards to PSs, no-fly cells and normal cells; we set ML parameters to values that we found them experimentally suitable and typically used in Q-learning projects \cite{b22} such as discount factor $\gamma = 0.9$, learning rate $\alpha = 0.1$, epsilon for greedy search $eps = 0.9/t$; and $N = 100000$ as the number of epochs that are used to be sure about the stability of results. The learning rate determines how much the new information overrides earlier information and how much of the earlier information will remain. In addition, Q-learning benefits from the greedy search through a random\_action function using epsilon, i.e. $eps$ variable as a holder for a given threshold value. If the random variable $eps$, is less than a threshold, in here set to $0.9/t$, then the random\_action function will randomly choose an action among all possible actions. Otherwise, the action that maximizes the total reward will be chosen. The epsilon greedy technique is called exploration and results in better discovery and understanding of the environment, especially in the beginning when the environment is unknown to the agent; and results in a better prediction and better policy. Therefore, we use a decaying epsilon that is high in the beginning i.e., when t is small but will lose its importance as the agent learns the environment; however, it will remain positive to model part of the small chance of stochastic nature of the system. 

The procedure shown in Algorithm 1 is a typical Q-learning algorithm with some small modifications. specifically considering rewards and values for the Start and Destination which are usually assumed to be zero. The role of epsilon in the greedy part of the algorithm is important in the beginning but diminishes as time passes and the agent learns the environment and then the actions will be chosen by an argmax function instead of random choices. The argmax function returns the action for which the maximum Q value is achieved. Applying the selected action, the next state and reward will be known and the Q value will be updated recursively. At the end of each iteration, the maximum Q value change is appended to deltas array to track the trend of diminishing changes towards convergence. Once the maximum change falls below a certain threshold, the iterations will end and the convergence is assumed. 

\begin{figure*}[htbp] 
    \centering
    \includegraphics[width=0.95\textwidth]{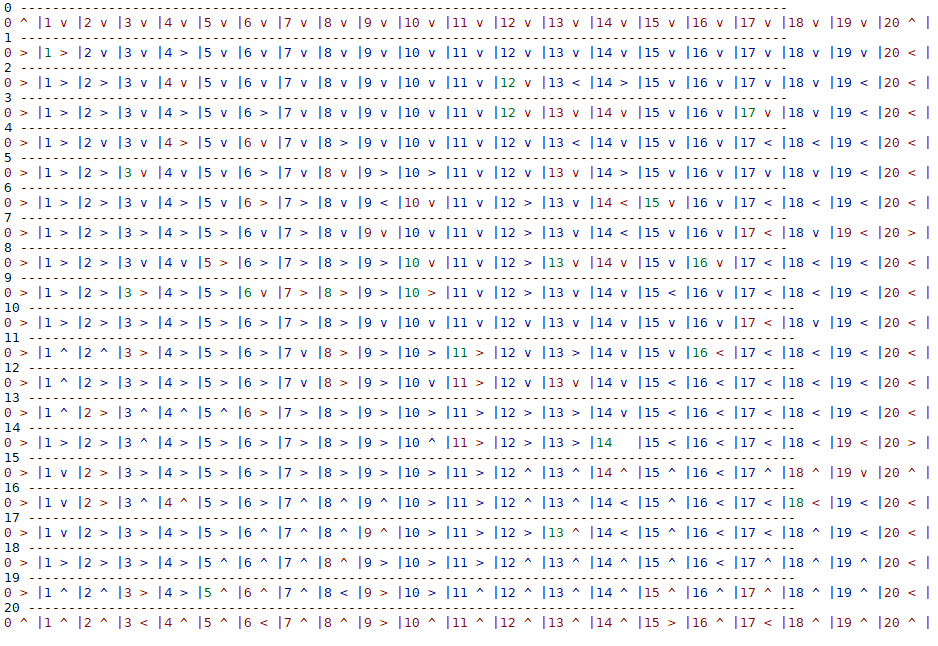}
     \caption{The map of trajectory of the UAV from each cell towards the Destination cell at (14, 14) via Up, Down, Right and Left moves.}
    \label{fig:f1}
\end{figure*}

\begin{figure}[htbp]
\centerline{\includegraphics[width=3.5in, height=2.5in]{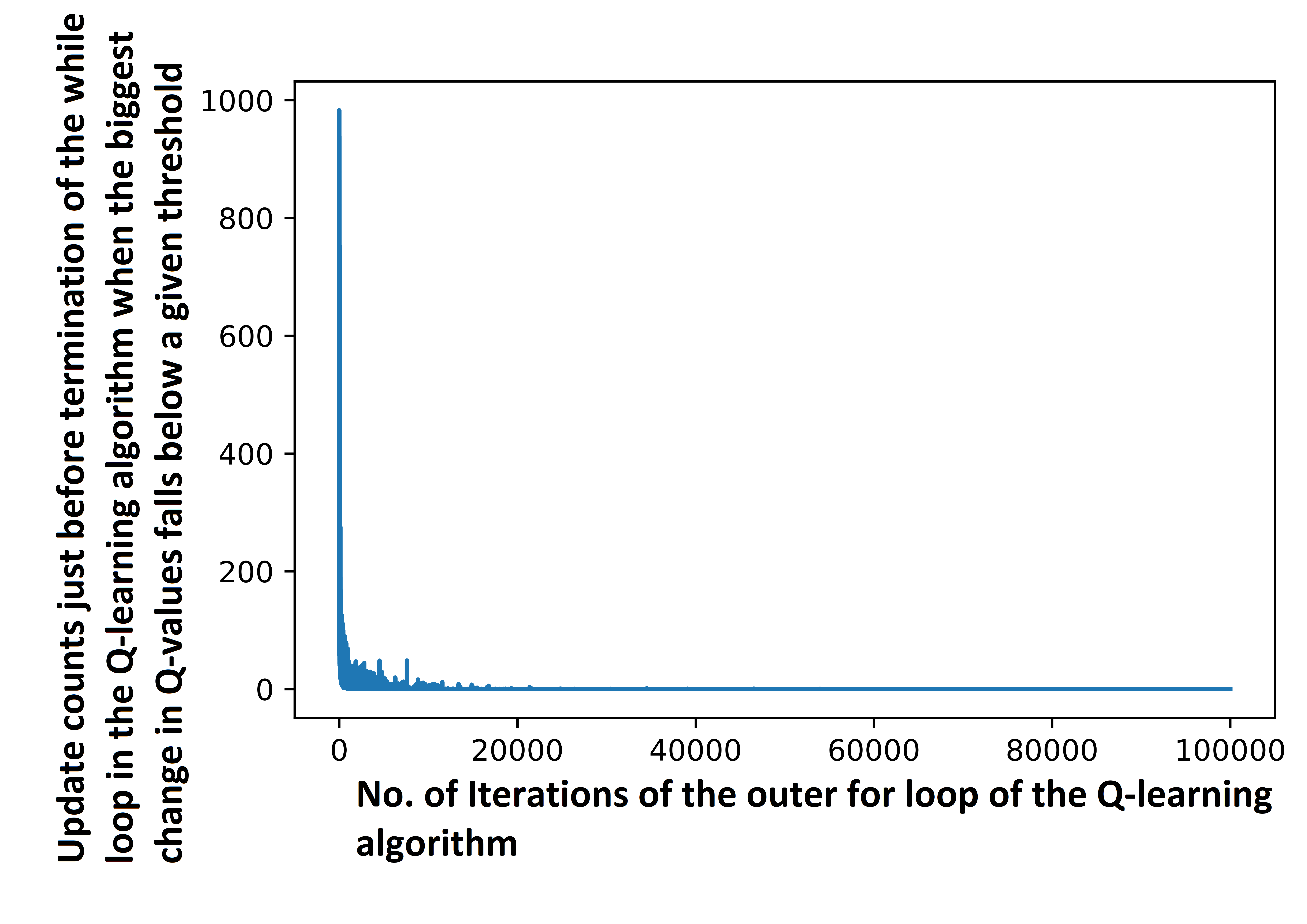}}
\caption{Q-learning convergence in UAV path optimization algorithm is achieved after about 20000 iterations when Q values are stabilized and the number of changes in Q values declines to 0.}
\label{fig:f2}
\end{figure}

\begin{figure}[htbp]
\centerline{\includegraphics[width=3.3in, height=2.9in]{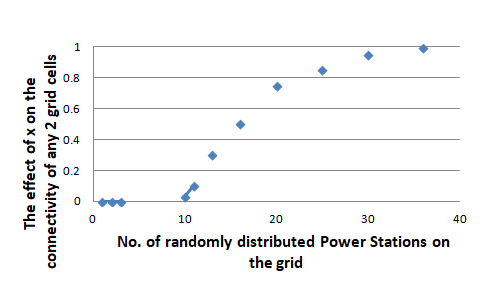}}
\caption{Connectivity rate between each two cells on the grid based on the number of PSs randomly distributed in the environment}
\label{fig:f3}
\end{figure}

\section{Results and Discussion}

Using Hierarchical RL can provide the UAV with a near optimal policy. \cite{b23} A typical result is shown in Fig. 2. This figure shows that the UAV learns to find its way to pass through the optimal path avoiding no-fly zones (column number indicated in red) while sweeping the maximum number of PS cells (column number indicated in green) without severe deviating from its shortest path mostly formed of regular cells (column number indicated in blue). In Fig. 2, letters "R", "L", "U" and "D" represent the action in the specific cell as moving right, left, up and down, respectively. Fig. 2 shows that any grid cell can be a starting cell but the Destination cell is fixed and in this particular example is considered to be the cell (14, 14). Our designated Start has been (1, 1). Therefore, in the settings depicted in Fig. 2, the UAV starts from (1,1) and goes one cell Down, then turns Right to the green PS cell (2,2) and so on until it will reach the Destination (14, 14). 


The high density of red cells near (14, 14) in Fig 2. and how the UAV manages to find its path to the Destination avoiding them show that the UAV could learn and extract complex plans in this area very well. The lowest battery level will correspond to the longest piece of the path between two PS cells i.e. just before the PS cell before which the UAV has flown the longest distance without recharge. Tracking the energy level of the scenario shown in Figure 2 shows that the lowest battery level in this scenario would occur at the Destination where the battery is only 30 percent full. The battery is more than 30\% full during this trajectory thanks to PSs at which it continually gets fully recharged.

A fact learned during training is that the notion of fairness implying justice matters even among things. To have a sustainable system among autonomous systems we need to create a model in which we assign relatively fair rewards and punishments in order that the system work properly. For example, if the UAV get little reward in PS cells compared to other cells, it may ignore that reward or if it receives an irrationally big reward in PS cells, it would prefers to settle around that cell and does not want to continue towards the Destination. On the other hand, receiving irrationally little punishment to no-fly cells results in ignoring it by the UAV passing through it while unfairly large punishment for no-fly cells impairs the path planning because it can cause avoiding the neighboring cells of that no-fly cell, sometimes even missing a useful PS in its neighborhood. Therefore, assigning relatively fair values of rewards and punishments is the key for correct performance of the system that can be achieved experimentally as we did in our experiments running the simulation program hundreds of times after a suitable initial setting learned from other papers. We fixed the key values by trial and errors as follows: the reward of each cell that has a PS, including the Start cell from which the UAV flies with 100\% charge, is set to be 1. The negative reward i.e. energy cost of each regular cell, that is neither PS nor no-fly, is set to be -0.1 that means a UAV after getting fully charged can fly away at most 10 steps before need to be recharged again. The big reward that is given when the UAV reaches the Destination is experimentally decided to be set to be 1000. Also, the negative reward of no-fly cells that should be very negative in order to make the UAV avoid them is experimentally set to be -30 after observing the simulation results in many times with different settings. Analysis of the repeatable successful results in all consecutive runs with no failure shows that the system with the current settings is promising because the UAV on the given grid network always works well and can find optimal paths independent of the number of PS cells and no-fly zones.   

Another issue that is worth attention is the problem of convergence in the RL. In around 1000 runs, our system has converged in all runs after a large number of iterations. For example, as illustrated in Fig. 3, the convergence is achieved after about $40000$ iterations, and despite rare small jumps due to reasons such as epsilon greedy learning having reached an acceptable stability. The convergence was monitored in each run of the code and the result was consistent with no evidence of opposite results. This result is intuitively predictable because after thousands of times repeating different trials and different routes in a limited system, the probability of unknown routes tends towards zero and even more importantly, the UAV finds nearly ideal path towards the Destination. However, there is always a small chance that the planned path is not the most optimal path, \cite{b23} but the UAV planner learns the map and its quality grows through large number of analyses and would improve its routes every time that it finds a new route better than the previous best route. However, in this type of problems, the curse of dimensionality matters and increasing the number of cells results in a sharp increase in number of iterations necessary in order to find a near optimal path.


The impact of the number of PSs on achieving a successful path through which the UAV reaches the Destination without any battery depletion from any starting grid cell was also analyzed and the results are illustrated in Fig. 4. For a 20*20 grid network with PSs that fully charge the UAV battery to value 1 and cell sizes that deplete 0.1 of its battery when moving from one cell to a neighboring cell, it is easy to see one can install, for example, 12 PSs and cover all 400 cells of the grid network provided that the PS locations were fixed and designed carefully and there were no no-fly zones imposed on the critical cells that are ideal locations for PSs. However, in the worst case scenario, we can imagine randomly distributed PSs. For example, if one imagine that PSs are also special UAVs that might change their places between charging the batteries of other UAVs. Therefore, we have also tried the analysis of randomly distributed PSs to see how the number of available PSs in our 20*20 grid network, will affect the likelihood of achieving a (near) optimal path by the UAV to reach the Destination without facing a dead battery. As seen in Fig. 4, the probability of safely traveling from the Start to the Destination starts to be greater than 0 once there are 10 extra PSs in the environment in addition to 2 PSs fixed in the Start and Destination. That's why 10 in Fig. 4 means 12 PSs in total and each number on the x axis shows the number of randomly distributed PSs in addition to those ones fixed in (1,1) and (14,14). The small initial probability in the beginning shows that the PSs must be located in special cells which is unlikely to happen in a random distribution, and the probability of successful missions will increase as the number of PSs increases and for 30 PSs and more randomly distributed in a 20*20 grid network, the probability of failure is almost 0.    


\section*{Conclusion}
The novelty of this work is that we have studied the path optimization of UAVs with respect to the energy constraint for the first time, in addition to other constraints such as connectivity and interference that were addressed by Challita et al. and Bulut et al. \cite{b12,b13}. We utilized Q-learning as a coarse planner successfully applied on top of others fine planners to simulate UAV's learning of an optimal path in long missions much beyond their battery capacity. For example, if interference matters in long missions, one can use our charge planner on top of the interference-aware planner of Chalita et al., or if the certain types of connectivity e.g. the one described by Bulut et al. matters, for long distances, one can use our energy planner on top of the connectivity planner of Bulut er al.  We have shown an example of a grid network with suitable sizes that is ready to be mounted on top of the planner pf Challita et al. and have shown that the UAV learns effectively to go from the Start large cell to the Destination large cell via optimal path i.e. a shortest feasible path via maximizing its benefit from PSs in the way and avoiding battery depletion, while other issues such as interference, connectivity, altitude, and other details can be managed by the fine RL-based planner such as that of Challita et al. We believe that this type of hierarchical planners is one of the best practical solutions and a right approach to solve the path optimization problems in automated vehicles such as self-driving cars and UAVs.

\vspace{12pt}




\begin{thebibliography}{00}
\bibitem{b1} A. E. Garcia et al., "Direct Air to Ground Communications for Flying Vehicles: Measurement and Scaling Study for 5G," 2019 IEEE 2nd 5G World Forum (5GWF), Dresden, Germany, 2019, pp. 310-315, doi: 10.1109/5GWF.2019.8911712.
\bibitem{b2} B. Van Der Bergh, A. Chiumento and S. Pollin, "LTE in the sky: trading off propagation benefits with interference costs for aerial nodes," in IEEE Communications Magazine, vol. 54, no. 5, pp. 44-50, May 2016, doi: 10.1109/MCOM.2016.7470934.
\bibitem{b3} X. Lin et al., "The Sky Is Not the Limit: LTE for Unmanned Aerial Vehicles," in IEEE Communications Magazine, vol. 56, no. 4, pp. 204-210, April 2018, doi: 10.1109/MCOM.2018.1700643. 
\bibitem{b4} M. M. Azari, F. Rosas, A. Chiumento and S. Pollin, "Coexistence of Terrestrial and Aerial Users in Cellular Networks," 2017 IEEE Globecom Workshops (GC Wkshps), Singapore, 2017, pp. 1-6, doi: 10.1109/GLOCOMW.2017.8269068.
\bibitem{b5} A. Fotouhi et al., "Survey on UAV Cellular Communications: Practical Aspects, Standardization Advancements, Regulation, and Security Challenges," in IEEE Communications Surveys \& Tutorials, vol. 21, no. 4, pp. 3417-3442, Fourthquarter 2019, doi: 10.1109/COMST.2019.2906228.
\bibitem{b6} A. Fotouhi, M. Ding and M. Hassan, "Flying Drone Base Stations for Macro Hotspots," in IEEE Access, vol. 6, pp. 19530-19539, 2018, doi: 10.1109/ACCESS.2018.2817799.
\bibitem{b7} A. Fotouhi, M. Ding and M. Hassan, "Service on Demand: Drone Base Stations Cruising in the Cellular Network," 2017 IEEE Globecom Workshops (GC Wkshps), Singapore, 2017, pp. 1-6, doi: 10.1109/GLOCOMW.2017.8269063.
\bibitem{b8} S. Zhang, Y. Zeng and R. Zhang, "Cellular-Enabled UAV Communication: A Connectivity-Constrained Trajectory Optimization Perspective," in IEEE Transactions on Communications, vol. 67, no. 3, pp. 2580-2604, March 2019, doi: 10.1109/TCOMM.2018.2880468.
\bibitem{b9} Y. Zeng, R. Zhang and T. J. Lim, "Throughput Maximization for UAV-Enabled Mobile Relaying Systems," in IEEE Transactions on Communications, vol. 64, no. 12, pp. 4983-4996, Dec. 2016, doi: 10.1109/TCOMM.2016.2611512.
\bibitem{b10} X. Li and J. Xu, "Positioning Optimization for Sum-Rate Maximization in UAV-Enabled Interference Channel," in IEEE Signal Processing Letters, vol. 26, no. 10, pp. 1466-1470, Oct. 2019, doi: 10.1109/LSP.2019.2934579.
\bibitem{b11} B. Galkin, J. Kibilda and L. A. DaSilva, "UAVs as Mobile Infrastructure: Addressing Battery Lifetime," in IEEE Communications Magazine, vol. 57, no. 6, pp. 132-137, June 2019, doi: 10.1109/MCOM.2019.1800545.

\bibitem{b12} U. Challita, W. Saad and C. Bettstetter, "Interference Management for Cellular-Connected UAVs: A Deep Reinforcement Learning Approach," in IEEE Transactions on Wireless Communications, vol. 18, no. 4, pp. 2125-2140, April 2019, doi: 10.1109/TWC.2019.2900035.
\bibitem{b13} E. Bulut and I. Guevenc, "Trajectory Optimization for Cellular-Connected UAVs with Disconnectivity Constraint," 2018 IEEE International Conference on Communications Workshops (ICC Workshops), Kansas City, MO, 2018, pp. 1-6, doi: 10.1109/ICCW.2018.8403623.
\bibitem{b14} S. Zhang, Y. Zeng and R. Zhang, "Cellular-Enabled UAV Communication: Trajectory Optimization under Connectivity Constraint," 2018 IEEE International Conference on Communications (ICC), Kansas City, MO, 2018, pp. 1-6, doi: 10.1109/ICC.2018.8422584.
\bibitem{b15} M. Chen, M. Mozaffari, W. Saad, C. Yin, M. Debbah and C. S. Hong, "Caching in the Sky: Proactive Deployment of Cache-Enabled Unmanned Aerial Vehicles for Optimized Quality-of-Experience," in IEEE Journal on Selected Areas in Communications, vol. 35, no. 5, pp. 1046-1061, May 2017, doi: 10.1109/JSAC.2017.2680898.
\bibitem{b16} B. Khamidehi and E. S. Sousa, "Reinforcement Learning-Based Trajectory Design for the Aerial Base Stations," 2019 IEEE 30th Annual International Symposium on Personal, Indoor and Mobile Radio Communications (PIMRC), Istanbul, Turkey, 2019, pp. 1-6, doi: 10.1109/PIMRC.2019.8904880.
\bibitem{b17} B. Khamidehi and E. S. Sousa, "A Double Q-Learning Approach for Navigation of Aerial Vehicles with Connectivity Constraint," ICC 2020 - 2020 IEEE International Conference on Communications (ICC), Dublin, Ireland, 2020, pp. 1-6, doi: 10.1109/ICC40277.2020.9148608.
\bibitem{b18} Watkins, Christopher J. C. H and Dayan, Peter, “Q-learning,” Machine learning, vol. 8, no. 3–4, pp. 279–292, 1992, doi: 10.1007/bf00992698.
\bibitem{b19} R. S. Sutton, A. G. Barto and R. J. Williams, "Reinforcement learning is direct adaptive optimal control," in IEEE Control Systems Magazine, vol. 12, no. 2, pp. 19-22, April 1992, doi: 10.1109/37.126844.
\bibitem{b20} N. K. Ure, G. Chowdhary, T. Toksoz, J. P. How, M. A. Vavrina and J. Vian, "An Automated Battery Management System to Enable Persistent Missions With Multiple Aerial Vehicles," in IEEE/ASME Transactions on Mechatronics, vol. 20, no. 1, pp. 275-286, Feb. 2015, doi: 10.1109/TMECH.2013.2294805.
\bibitem{b21} M. Bekhti, M. Abdennebi, N. Achir and K. Boussetta, "Path planning of unmanned aerial vehicles with terrestrial wireless network tracking," 2016 Wireless Days (WD), Toulouse, 2016, pp. 1-6, doi: 10.1109/WD.2016.7461521.
\bibitem{b22} T. Okuyama, T. Gonsalves and J. Upadhay, "Autonomous Driving System based on Deep Q Learnig," 2018 International Conference on Intelligent Autonomous Systems (ICoIAS), Singapore, 2018, pp. 201-205, doi: 10.1109/ICoIAS.2018.8494053.
\bibitem{b23} Nachum, Ofir, Gu, Shixiang, Lee, Honglak, and Levine, Sergey, “Near-Optimal Representation Learning for Hierarchical Reinforcement Learning,” International Conference on Learning Representations (ICLR), 2018. arXiv preprint arXiv:1810.01257.
\end{thebibliography}
\end{document}